\documentclass{article}

\usepackage{microtype}
\usepackage{subfigure}
\usepackage{booktabs} 
\usepackage{bbm}
\usepackage{mathtools}
\usepackage[numbers]{natbib}
\usepackage{verbatim}
\usepackage{enumitem}
\setlist[itemize]{topsep=2pt, leftmargin=5mm}
\usepackage{wrapfig}

\usepackage{url}
\usepackage{multirow}
\usepackage{makecell}
\usepackage{algorithm}
\usepackage{algpseudocode}
\algnewcommand{\Initialize}[1]{%
  \State \textbf{Initialize:}
  \Statex \hspace*{\algorithmicindent}\parbox[t]{.8\linewidth}{\raggedright #1}
}
\usepackage[table,x11names]{xcolor}
\usepackage{collcell}
\usepackage{array}
\usepackage{tikz}
\usetikzlibrary{backgrounds,fit}
\usetikzlibrary{shapes.geometric}
\usetikzlibrary{decorations.pathreplacing}

\usepackage{pgfkeys}
\usepackage{graphicx}
\usepackage{changepage}
  \pgfkeys{/heat/.is family, /heat,
      Max colour/.initial = Green4,
      Min colour/.initial = Red1,
      max colour/.initial = SpringGreen3,
      min colour/.initial = Yellow1,
      text colour/.initial = black,
      Min color/.style = {Min colour=#1},
      Max color/.style = {Max colour=#1},
      min color/.style = {min colour=#1},
      max color/.style = {max colour=#1},
      text color/.style = {text colour=#1},
      min/.initial = -1,
      max/.initial = 1,
      slider/.code={%
         \tikz{\shade[left color=\HVal{min colour},%
                      right color=\HVal{max colour}]%
            (current page.south west) rectangle ++(#1,12pt);
         }%
      }%
  }
  \newcommand\Heatset[1]{\pgfkeys{/heat, #1}}
  \newcommand\HVal[1]{\pgfkeysvalueof{/heat/#1}}
  \newcolumntype{H}{>{\collectcell\Heat}r<{\endcollectcell}}
  \newcommand\Heat[1]{
    \if\relax\detokenize{#1}\relax
    \else%
      \pgfmathparse{int(100*(#1-\HVal{min})/(\HVal{max}-\HVal{min}))}
      \ifnum\pgfmathresult>100
        \edef\HeatCell{\noexpand\cellcolor{\HVal{Max colour}}}%
      \else\ifnum\pgfmathresult<0
          \edef\HeatCell{\noexpand\cellcolor{\HVal{Min colour}}}%
        \else
          \edef\HeatCell{\noexpand\cellcolor{\HVal{max colour}!\pgfmathresult!\HVal{min colour}}}%
        \fi%
      \fi%
      \HeatCell\textcolor{\HVal{text colour}}{$#1$}%
    \fi%
  }
\newcommand{\cmmnt}[1]{}

\usepackage{todonotes}

\usepackage{microtype}
\usepackage{subfigure}
\usepackage{booktabs} 
\usepackage{wrapfig}

\usepackage{amsthm,amsmath,amssymb, amsfonts}

\allowdisplaybreaks

\DeclareMathOperator*{\mode}{Mode}

\usepackage[preprint]{neurips_2021}
\usepackage{comment}

\title{On-the-fly Strategy Adaptation for ad-hoc Agent Coordination}

\author{
  Jaleh Zand \\
  University of Oxford\\
  \texttt{jz@robots.ox.ac.uk} \\
  \And
  Jack Parker-Holder \\
  University of Oxford \\
   \texttt{jackph@robots.ox.ac.uk} \\
  \AND
  Stephen J. Roberts \\
  University of Oxford \\
  \texttt{sjrob@robots.ox.ac.uk } \\
   }

\begin{document}

\maketitle

\begin{abstract}
Training agents in cooperative settings offers the promise of AI agents able to interact effectively with humans (and other agents) in the real world. Multi-agent reinforcement learning (MARL) has the potential to achieve this goal, demonstrating success in a series of challenging problems. However, whilst these advances are significant, the vast majority of focus has been on the \emph{self-play} paradigm. This often results in a \emph{coordination problem}, caused by agents learning to make use of arbitrary conventions when playing with themselves. This means that even the strongest self-play agents may have very low \emph{cross-play} with other agents, including other initializations of the same algorithm. In this paper we propose to solve this problem by adapting agent strategies \emph{on the fly}, using a posterior belief over the other agents' strategy. Concretely, we consider the problem of selecting a strategy from a finite set of previously trained agents, to play with an unknown partner. We propose an extension of the classic statistical technique, Gibbs sampling, to update beliefs about other agents and obtain close to optimal ad-hoc performance. Despite its simplicity, our method is able to achieve strong cross-play with unseen partners in the challenging card game of Hanabi, achieving successful ad-hoc coordination without knowledge of the partner's strategy a priori. \end{abstract}

\section{Introduction}
In recent years there has been a surge of interest in deep multi-agent reinforcement learning (MARL, \cite{foerster2016learning, foersterthesis}), a paradigm whereby agents learn, from interaction, to compete or cooperate with other agents. This paper focuses on cooperative MARL, which not only offers the potential for agents to learn to interact with other machine learning based agents in the real world, but also provides a possible framework to coordinate with humans.

Many of the most prominent successes in MARL have come in the \emph{self-play} paradigm \cite{tesauro1994td} where agents optimise their policies by playing a copies of themselves repeatedly without any direct supervision. The self-play setting has been used in both adversarial and cooperative MARL settings. There has been great success with a series of agents achieving breakthroughs on challenging environments such as Go \cite{silver2018general, silver2017mastering}, Chess \citep{silver2018general, silver2017chess}, Poker \cite{moravvcik2017deepstack, brown2018superhuman} and Hanabi \cite{bard2020hanabi, hu2019simplified, foerster2019bayesian}.

It is shown that in a two-player game, a self-play strategy could converge to a Nash equilibrium \cite{nash20164} and could even lead to super-human performance in certain domains \cite{zinkevich2007regret, bai2020provable, crandall2011learning, conitzer2007awesome}. However the self-play framework seems more beneficial in an adversarial MARL setting compared to a cooperative one. For one thing, \emph{regret} is not easily defined in a self-play setting. In adversarial zero-sum games the strongest regret is considered by using a more tractable notion of \emph{external regret} \cite{blum2007external, arora2012online}. Such relaxation however is weak in a cooperative setting \cite{digiovanni2021survey} and overall regret minimization is not clearly defined and understood in a non zero-sum-game or an $n$-player setting. In the cooperative MARL framework  such as Hanabi for example, despite the strength of these agents when playing with copies of themselves, they often achieve this performance by making use of arbitrary conventions. For example, in Hanabi, agents will play hint tokens in a specific manner, only interpretable to themselves \cite{hu2020other}. This means many state-of-the-art agents fail to coordinate with an unknown, arbitrary agent at test time \cite{bard2020hanabi}, achieving significantly worse \emph{cross-play} compared to their promising self-play performance. Furthermore, many prominent MARL agents fail to coordinate with independent runs of the same agent algorithm \cite{hu2020other}, suggesting that multiple initializations of the same agent converge to different strategies (described as the ad-hoc coordination problem).

In this paper we consider the \emph{ad-hoc coordination problem} \cite{adhoc_stone}, whereby an agent has to coordinate with an unknown partner in either an ad-hoc paradigm or with just a few successive trials. We focus on the problem of selecting a strategy from a finite set of previously trained agents and interacting with an agent whose policy is unknown. In this setting the agents need to actively interact and make inferences about other agents to achieve successful cooperation. Humans do this routinely in their interactions as suggested by the theory of mind \cite{baker2017rational,premack1978does}.

At the core of our approach we use Gibbs sampling \cite{gemanstochastic}. Although Gibbs sampling is well-known, as far as we are aware it has not previously been considered for ad-hoc coordination in multi-agent settings. The method is prominently used for Bayesian inference, and has been applied in various areas of machine learning; some recent work includes \cite{ludwig2020database, ko2019accelerating, ko2019flexgibbs, de2018minibatch} in probabilistic graphical models and \cite{alt2019correlation, yu2019multi} in reinforcement learning. The method forms the core of inference in many probabilistic programming languages such as OpenBUGS \cite{spiegelhalter2007openbugs}, JAGS \cite{plummer2003jags}, Church \cite{paige2014compilation}, PyMC \cite{patil2010pymc}, and Turing \cite{ge2018t}. Gibbs sampling is a simulation approach to estimate the joint probability distribution by sampling from the conditional probability distributions. By iteratively updating conditional probabilities, thus evaluating beliefs about the other agents' strategies and hidden information, agents are able to form an understanding of the other agents' policies in such a way that they can perform close to optimal performance, subject to their capabilities, even in an ad-hoc environment. Despite its simplicity, our method is able to achieve strong cross-play with unseen partners in the challenging card game of Hanabi, achieving successful ad-hoc coordination.   

While many agents can achieve impressive self-play performance, they often do so by developing arbitrary conventions which hinder their effectiveness when coordinating with others. A key challenge in cooperative multi-agent RL is the ability to train robust agents that can play well with others, including humans \cite{DBLP:journals/corr/abs-2012-08630}. We focus on the multi-player card game of Hanabi \cite{bard2020hanabi}, being one of the most prominent testbeds for cooperative MARL \cite{foerster2019bayesian, lerer2019improving, hu2019simplified, hu2020other, nekoei2021continuous}. Hanabi is a partially-observable, fully cooperative multi-agent game that consists of 2-5 players. What makes it such a fascinating testbed for MARL is the rich volume of potential strategies. For instance, agents may use completely different conventions, thus two agents may infer different beliefs around the same hint or action taken. The game therefore represents a multi-challenge problem where one can employ the theory of mind. One of the prominent challenges in the Hanabi environment is \emph{ad-hoc team play}, the ability to play well with an arbitrary player never seen before. As of yet, very few agents have been able to achieve strong performance in this setting, and those that have rely on significant domain knowledge \cite{offbelieflearning}.

The contributions of this paper are as follows. First, to the best of our knowledge, there is no prior work that considers Gibbs sampling in cooperative multi-agent settings. Second, we show that, using an extension of the Gibbs algorithm, we are able to achieve close to self-play performance in cross-play situations with ad-hoc players. Thirdly, while there have been many works that consider coordination amongst ad-hoc agents, most of these are limited to agents that are trained by the same model,
or are not constrained to a single-shot ad-hoc setting. Our method makes comparisons across a set of diversified models and therefore a diverse set of ad-hoc agents. Finally, since there are no restrictions on the policies that can be included, our method offers scope to be applied to future, stronger base policies.

\section{Related Work} \label{sec:related}
Bowling and McCracken \cite{bowling2005coordination}, and Stone et al. \cite{adhoc_stone} were among the first to propose the \emph{ad-hoc} teamwork challenge, whereby agents need to coordinate with a previously unknown agent in a task where all agents are capable of individually contributing. More recently, the authors of the Hanabi challenge \cite{bard2020hanabi} explicitly propose \emph{ad-hoc team play} as one of the primary benchmarks for future progress in cooperative MARL. Concretely, the challenge states that an agent must achieve a high score in the game when coordinating with an unknown partner, with just a few ``warm-up'' games. The most related work to ours is \cite{mapelites_hanabi} who use the MAP-Elites \cite{Mouret2015IlluminatingSS} algorithm to learn a diverse set of policies, and show it can be used to coordinate across seeds of the same algorithm by making use of an iterative Bayesian Optimization approach \cite{bayesopt_nando}. However, this approach requires meta-information about the policies in the set, which requires human knowledge and thus does not generalize to other problems.

In other related work in ad-hoc multi agent problems, the MARL agent has access to other agent's previous observed behaviour \cite{barrett2017making, peysakhovich2017prosocial, lerer2019learning, carroll2019utility} or it is assumed that the agents will interact with the same agents interacted with during training \cite{lowe2017multi}. It has also been shown it is possible to solve various MARL tasks by learning communication conventions between agents using communication channels \cite{sukhbaatar2016learning, mordatch2018emergence}. In our setting the agents have no access to previous behaviour or observations and the are no channels to establish communication conventions.

Related is the recently introduced \emph{zero-shot coordination} (ZSC) problem \cite{hu2020other}. ZSC specifies that an algorithm must be run independently and produce agents with high cross-play performance. Methods for achieving this include making use of the symmetries of the problem \cite{hu2020other, ridgerider}, or making assumptions about prior actions \cite{offbelieflearning}. ZSC differs from the ad-hoc setting in that it requires agents to train from scratch and produce reproducible policies \emph{within the same algorithm}, but it does not place any requirements on these subsequent agents to play well with unknown agents such as humans. Nonetheless, methods for ZSC have produced more general agents that have successfully transferred to the ad-hoc teamplay setting \cite{offbelieflearning}. Finally, another recent line of work considers training a best response to a diverse population of agents for robust zero-shot and ad hoc coordination \cite{strouse2021fcp}. This is in contrast to OSA which instead seeks to select a single agent from the pre-trained population.

\section{Background}
\label{sec:background}
In this paper we consider fully-cooperative Markov games. We model this setting with a Decentralized Partially-Observable Markov Decision Process (Dec-POMDP~\cite{bernstein2002complexity,nair2003taming}), defined as a tuple $G=\{\mathcal{S}, \mathcal{U}, P, R,\Omega, O, \gamma\}$, with states $s_t \in \mathcal{S}$. There are $i=1, \dots, N$ agents who each choose actions, $u^i_t \in \mathcal{U}$ at each time step, in which $\Omega$ is the the set of observations for each agent. The game is partially observable, with $o^i_t \sim O(o|i, s_t)$ defined as each agent's (stochastic) observation function. At time $t$ each agent has an action-observation history $\tau^i_t= \{o^i_0, u^i_0, r^i_0, \cdots, o^i_t\}$ and selects action $u^i_t$ using stochastic policies of the form $\pi^i_\theta(u^i|\tau^i_t)$.  The transition function, $P(s'|s,\mathbf{u})$, is conditioned on the joint action, $\mathbf{u}$.

The game is fully cooperative, thus agents share the reward $r_t = R(s_t,u_t)$, conditioned on the joint action and the state. The goal is hence to maximize the expected return, $J = \mathbb{E}_\tau R(\tau)$, where $ R(\tau) =\sum_t \gamma^t r_t$ is calculated using a discount factor $\gamma$.

As in Foerster et al. \cite{foerster2019bayesian}, we consider a setting in which the Markov state, $s_t$, consists of discrete features, $f^{s_t}$, which are themselves composed of public features, $f^{\text{pub},s_t}$, and private features, $f^{\text{pri},s_t}$. The public features are known by all agents and the private features are known by at least one, but not all, the agents. We denote $f^{i,s_t}$ as representing the private features only observed by agent $i$. In addition, we assume two types of agents in the system; simple agents, $A^S$, that play a fixed policy, $\pi^S$, and complex agents, $A^C$, which coordinate with the set of simple agents and further have access to the set of policies, $\Pi = \{\pi^1,...,\pi^n\}$. This provides groups of agents with the ability to reason over the intention of others, similar to the \emph{theory of mind} popularized for understanding human action taking \cite{baker2017rational,premack1978does}.

Similar to the approach in \cite{bard2013online}, the portfolio of the policies, $\Pi$, is constructed offline. The portfolio consists of a set of policies and for each policy $\pi^i \in \Pi$ played by the other agents, the optimal response policy is $B(\pi^i)$. In this work the policy set $\Pi$, provided to our agent, is not optimised. This could be an area of focus for future work. Further, the function $B(\pi^i)$ maps $\pi^i \in \Pi$ to $\pi^j \in \Pi$. In a more complex setting this assumption could be relaxed, such that instead of $\Pi$ there is a partner policy set and a response policy set that do not overlap.

\section{On-the-fly Strategy Adaptation}
In this section we briefly overview Gibbs sampling and detail its selection as a credible method for strategy selection in ad-hoc agent settings. We then introduce our algorithm, On-the-fly Strategy Adaptation (OSA).

\subsection{Gibbs Sampling} \label{sec:Gibbs_sampling}

In line with the \emph{theory of mind}, one possible approach for coordination in  multi-agent ad-hoc problems is the formation of beliefs regarding strategies that are played by the other agents in the system. If agents can accurately identify and understand the strategies that are played by other agents, they can respond with a policy that could achieve the best results.  In a partially-observable, multi-agent problem space however, agents need to estimate the joint probability distribution of the other agents' policies and the hidden information, which, by definition, is not available to them. This joint probability distribution is more often than not difficult to infer and thence sample from.

In our ad-hoc coordination settings, the complex agents, $A^C$, need to estimate the joint probability distribution $P(\pi^S,f^{S,s_t} \mid u_t^S, s_t)$ at every step of the game, where $\pi^S$ is the simple agent policy, and $f^{S,s_t} \in F^{S,s_t} = \{f^{1,s_t},...,f^{n,s_t}\}$. Here, $F^{S,s_t}$ denotes a set of features that $A^S$ could be privy to but are hidden from $A^C$ in state $s_t$. For example, in a typical card game, $F^{S,s_t}$ could represent a combination of cards at each step that only $A^S$ can observe and are not openly discarded or on display.

\begin{algorithm}
  \caption{Gibbs Sampling}
  \label{Gibbs_algo}
  \begin{algorithmic}[1]
  \Initialize{\strut$ \pi_0^S \gets \pi^i$, $\pi^i \in \Pi = \{\pi^1,\ldots,\pi^n\}$ }
      \For{$t = 0,...,T$}
            \If {it is $A^{C}$ turn to play}
            \State Sample $f_{t+1}^{S,s_t} \sim P(f^{S,s_t} \mid \pi_t^S, u_t^S, s_t)$
            \State Sample ${\pi}^S_{t+1} \sim P( \pi^S \mid f_{t+1}^{S,s_t}, u_t^S, s_t) $
		   
      \EndIf
      \EndFor
  \end{algorithmic}
\end{algorithm}

In order to estimate this joint distribution we utilize an extension of the Gibbs sampling algorithm \cite{bishop}. Gibbs sampling facilitates the estimation of a joint distribution by sampling from the conditional probabilities of that distribution. Gibbs sampling is a useful method when the full joint distribution is intractable or cannot be evaluated or sampled from directly, but the conditional distribution of each variable (or clique of variables) can be evaluated and hence sampled from. Being able to estimate complex joint distributions from sequences of samples from the conditionals, makes Gibbs sampling a powerful approach to solutions to the joint probability distribution in ad-hoc coordination problems. However, Gibbs sampling in its standard form can be slow to converge and can only provide per-step evaluations in our problem domain. We therefore develop an extended version of the Gibbs approach which is further discussed in Section \ref{sec:osa}. The core of the two-step Gibbs sampling algorithm we use here is shown in Algorithm \ref{Gibbs_algo}.

Noting that all agents observe $u^S_t$ at every step, we may use Bayes' theorem to estimate the distributions in steps 4 and 5 of Algorithm \ref{Gibbs_algo} as follows:

\begin{align}
    P(f^{S,s_t} | \pi^S_t, u_t^S, s_t)  
    = \dfrac{P(u_t^S | f^{S,s_t}, \pi^S_t, s_t) P(f^{S,s_t} \mid \pi^S_t, s_t)} {P(u_t^S | \pi^S_t, s_t)}\\
    = \dfrac{P(u_t^S | f^{S,s_t}, \pi^S_t, s_t) P(f^{S,s_t}\mid \pi^S_t, s_t )}{\sum_{f^{i,s_t}} P(u_t^S | f^{i,s_t},\pi^S_t, s_t)P(f^{i,s_t} \mid \pi^S_t, s_t )}  \label{eq:f}
\end{align}

\begin{align}
        P( \pi^S | f^{S,s_t}_{t+1}, u_t^S, s_t)
    = \dfrac{P(u^S_t | \pi^S, f^{S,s_t}_{t+1}, s_t)P(\pi^S \mid f^{S,s_t}_{t+1}, s_t)}{P(u_t^S | f^{S,s_t}_{t+1},s_t)}\\
    = \dfrac{P(u^S_t | \pi^S, f^{S,s_t}_{t+1}, s_t)P(\pi^S \mid f^{S,s_t}_{t+1}, s_t)}{\sum_{\pi^i} P(u_t^S | {\pi^i},f^{S,s_t}_{t+1},s_t)P(\pi^i \mid f^{S,s_t}_{t+1}, s_t)} \label{eq:pi}
\end{align}
\vspace{0mm}

We further note that the Gibbs approach facilitates a per-step evaluation of $\pi^S$ for the complex agents. This, however, is not sufficient to accurately estimate the policy for $A^S$. We use the most frequent value of the per step evaluations and this is further discussed in Section \ref{sec:osa}.

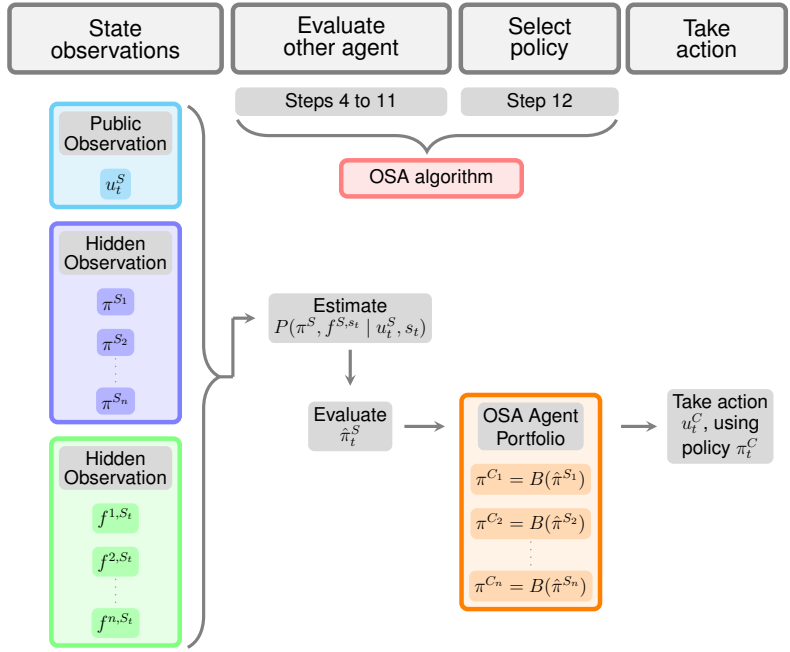
\begin{figure}[htb]
\vspace{2mm}
\centering
\scalebox{0.6}{
\tikzstyle{dis}=[loosely dotted,minimum size = 1.2cm, thick, draw =gray!80]
\tikzstyle{drg} = [-stealth, thick, gray!100, line width=2pt]
\tikzstyle{drg_l} = [thick, gray!100, line width=2pt]

\begin{tikzpicture}[outer sep=0.05cm,node distance=0.8cm]

\tikzstyle{bigbox} = [draw=blue!50, line width=1mm, fill=blue!10, rounded corners, rectangle, minimum size = 1.4cm]
\tikzstyle{box} = [font=\sffamily,minimum size=0.6cm, rounded corners,rectangle, fill=blue!30]

\tikzstyle{bigbox_g} = [font=\sffamily,draw=green!50, line width=1mm, fill=green!10, rounded corners, rectangle, minimum size = 1.4cm]
\tikzstyle{box_g} = [font=\sffamily,minimum size=0.6cm, rounded corners,rectangle, fill=green!30]

\tikzstyle{bigbox_or} = [font=\sffamily,draw=cyan!50, line width=1mm, fill=cyan!10, rounded corners, rectangle, minimum size=1.6cm]
\tikzstyle{box_or} = [font=\sffamily, minimum size=0.6cm, rounded corners,rectangle, fill=cyan!30]

\tikzstyle{bigbox_y} = [font=\sffamily,draw=orange!100, line width=1mm, fill=orange!10, rounded corners, rectangle, minimum size=2cm]
\tikzstyle{box_y} = [font=\sffamily, minimum size=0.6cm, rounded corners,rectangle, fill=orange!30]

\tikzstyle{box_gr} = [font=\sffamily, minimum size=0.6cm, rounded corners,rectangle, fill=gray!30]

\tikzstyle{box_gray} = [font=\sffamily, draw=gray!100, line width=1mm, fill=gray!10, rounded corners, rectangle, minimum size=1cm, minimum width=3.5cm, minimum height=1.5cm]
\tikzstyle{box_grayl} = [font=\sffamily, draw=gray!100, line width=1mm, fill=gray!10, rounded corners, rectangle, minimum size=1cm, minimum width=4.8cm, minimum height=1.5cm]
\tikzstyle{box_graym} = [font=\sffamily, draw=gray!100, line width=1mm, fill=gray!10, rounded corners, rectangle, minimum size=1cm, minimum width=4.7cm, minimum height=1.5cm]

\tikzstyle{box_red} = [font=\sffamily, draw=red!50, line width=1mm, fill=red!10, rounded corners, rectangle, minimum size=1cm, minimum width=4cm, minimum height=0.8cm]

\node[box_gr] (100) {\large \shortstack{Public \\ Observation}};
\node[box_or,below=0.2cm of 100] (101) {\large $u^S_t$};
\begin{pgfonlayer}{background}
  \node[bigbox_or] [fit = (100) (101)] {};
\end{pgfonlayer}
\node[box_gr,below=0.6cm of 101] (0) {\large \shortstack{Hidden \\ Observation}};
\node[box,below=0.2cm of 0] (1) {\large $\pi^{S_1}$};
\node[box,below=0.2cm of 1] (2) {\large $\pi^{S_2}$};
\node[box,below =0.6cm of 2] (3) {\large $\pi^{S_n}$};
\draw[dis] (2) to (3);
\begin{pgfonlayer}{background}
  \node[bigbox] [fit = (0) (3)] {};
\end{pgfonlayer}

\node[box_gr,below=0.6cm of 3] (10) {\large \shortstack{Hidden \\ Observation}};
\node[box_g,below=0.2cm of 10] (4) {\large $f^{1,S_t}$};
\node[box_g,below=0.2cm of 4] (5) {\large $f^{2,S_t}$};
\node[box_g,below =0.6cm of 5] (6) {\large $f^{n,S_t}$};
\begin{pgfonlayer}{background}
  \node[bigbox_g] [fit = (10) (6)] {};
\end{pgfonlayer}
\draw[dis] (5) to (6);

\draw [decorate,decoration={brace,amplitude=20pt, mirror},xshift=0pt,yshift=-9.7cm, gray!100,line width=2pt] (1.6,-1.7) -- (1.6,10.3) node [gray!80,midway,xshift=-0.6cm] {};

\draw[drg_l] (2.3,-5.4) -- (2.6,-5.4);
\draw[drg_l] (2.6,-5.4) -- (2.6,-4.1);
\draw[drg] (2.6,-4.1) -- (3.1,-4.1);

\node[box_gr] at (5.2, -4.1)   (a) {\large \shortstack{Estimate \\ $P(\pi^S,f^{S,s_t} \mid u_t^S, s_t)$}};
\draw[drg] (5.2,-4.8) -- (5.2,-5.6);

\node[box_gr] at (5.2, -6.5)   (a) {\large \shortstack{Evaluate \\ $\hat{\pi}^S_t$}};
\draw[drg] (6.4,-6.5) -- (7.3,-6.5);

\node[box_gr] at (9.2, -6.5)   (20) {\large \shortstack{OSA Agent \\ Portfolio}};
\node[box_y,below=0.2cm of 20] (21) {\large $\pi^{C_1} = B(\hat{\pi}^{S_1})$};
\node[box_y,below=0.2cm of 21] (22) {\large $\pi^{C_2} = B(\hat{\pi}^{S_2})$};
\node[box_y,below=0.6cm of 22] (23) {\large $\pi^{C_n} = B(\hat{\pi}^{S_n})$};
\begin{pgfonlayer}{background}
  \node[bigbox_y] [fit = (20) (23)] {};
\end{pgfonlayer}
\draw[dis] (22) to (23);

\draw[drg] (11.2,-6.5) -- (12,-6.5);
\node[box_gr] at (13.4, -6.5)   (b) {\large \shortstack{Take action \\ $u^C_t$, using \\ policy $\pi^C_t$}};

\node[box_graym] at (0, 2.1)   (b1) {\Large \shortstack {State \\ observations} };

\node[box_grayl] at (5, 2.1)   (b2) {\Large \shortstack{Evaluate \\ other agent} };

\node[box_gray] at (9.4, 2.1)   (b3) {\Large \shortstack{Select \\ policy} };

\node[box_gray] at (13.1, 2.1)   (b4) {\Large \shortstack{Take \\ action} };

\node[font=\sffamily, minimum size=0.6cm, rounded corners,rectangle, fill=gray!30, minimum width=4.7cm] at (5, 0.7)   (b4) {\large \shortstack{Steps 4 to 11} };
\node[font=\sffamily, minimum size=0.6cm, rounded corners,rectangle, fill=gray!30, minimum width=3.5cm] at (9.4, 0.7)   (b4) {\large \shortstack{Step 12} };

\draw [decorate,decoration={brace,amplitude=20pt, mirror},xshift=0pt,yshift=-9.7cm, gray!100,line width=2pt] (2.7,10) -- (11.1,10) node [gray!80,midway,xshift=-0.6cm] {};

\node[box_red] at (7, -1)   (b4) {\large OSA algorithm };

\end{tikzpicture}
}
\vspace{2mm}
\caption{\textit{Schematic of the steps of the OSA algorithm at each time step}}
\label{fig:schematic}

\end{figure}

\subsection{The OSA Algorithm}\label{sec:osa}

We introduce On-the-fly Strategy Adaptation (OSA) as an extension to Algorithm \ref{Gibbs_algo}. In the OSA approach, the set of $A^C$ perform the Gibbs update steps of Algorithm \ref{Gibbs_algo} to evaluate $\pi^S_t$, providing per-step evaluation of $\pi^S$. Each agent in $A^C$ then uses the most-likely policy evaluation, $\hat{\pi}^S_t$, as its policy belief over $A^S$ at time $t$. We note therefore, that $\hat{\pi}^S_t = \mode(\pi^S_1,...,\pi^S_{t+1})$. Agents $A^C$ then set their policy $\pi^C_t$ to $B(\hat{\pi}^S_t)$, where $B(\hat{\pi}^S_t)$ is the optimal response policy for $\hat{\pi}^S_t$.

In addition to $A^C$ evaluating each policy in $\Pi$ based on the observed action $u^S_t$, we augment the approach to remove redundant policies from the set. For every policy $\pi^i \in \Pi$, if $P( u_t^S \mid \pi^i, s_t) \approx 0$, we remove that policy from the policy set (Algorithm \ref{osa_algo}, steps 6-8). This improves accuracy when $\Pi$ is a large set and reduces the computational overheads of the approach by removing poor policies early on.

The full OSA procedure is shown in Algorithm \ref{osa_algo}. We use Equations \ref{eq:f} and \ref{eq:pi} to evaluate steps 4 and 10 in the algorithm. Figure \ref{fig:schematic} presents the schematics of the algorithm.


\begin{algorithm}[htb]
  \caption{OSA algorithm}\label{osa_algo}
  \hspace*{\algorithmicindent} \textbf{Input}: set of fixed policies $\Pi = \{{\pi}^1,...,{\pi}^n\}$\
  \begin{algorithmic}[1]
  \Initialize{\strut$ \hat{\pi}_0^S \gets \pi^i$, $\pi^i \in \Pi = \{{\pi}^1,...,{\pi}^n\}$ 
  }
      \While{the game is ongoing}
            \If {it is $A^{C}$ turn to play}
            \State $f^{S,s_t}_{t+1} \sim P(f^{S,s_t} \mid \hat{\pi}_t^S, u_t^S, s_t)$
            \For{$\pi^i \in \{\pi^1,\ldots,\pi^n\}$}
                \If {$P( u_t^S \mid \pi^i, s_t) \approx 0$} 
                \State $\Pi = \Pi \setminus \{\pi^i\}$ 
                \Comment{Remove redundant policies}
                \EndIf
            \EndFor
            \State $\pi^S_{t+1} \sim P( \pi^S \mid f^{S,s_t}_{t+1}, u_t^S, s_t) $
		    \State $\hat{\pi}^S_{t+1} = \mode(\pi^S_1,...,\pi^S_{t+1})$
		    \Comment{$\hat{\pi}^S_{t}$ is the most frequent $\pi^S_t$}
		    \State $\pi_{t+1}^C = B(\hat{\pi}^S_{t+1}) $
		    \Comment{$B$: optimal response policy function}
      \EndIf
      \EndWhile
  \end{algorithmic}
  \vspace{0mm}
\end{algorithm}



\section{Experiments} \label{sec:experiments}
In this section we test the ability of OSA to find effective ad-hoc coordination policies. To begin, we consider a toy setting, where agents must coordinate in a simple game of volleyball. We then consider the challenging card game of Hanabi, one of the most prominent testbeds for cooperative AI, where we seek to achieve coordination between a set of MARL and hand coded agents.

For all evaluations presented in this paper, we used Intel Xeon Gold CPUs (with 48 cores), running at 2.3GHz and with 256 GB RAM.

In our experiments, we assume that all policies in $\Pi$ are equally likely for $A^C$ to interact with and therefore $A^C$ assumes a uniform prior distribution over $\pi^i \in \Pi$. We note that this assumption is not necessary for the algorithm and can be relaxed. Further in our first experiment, the Slime Volleyball game $F^{S,s_t}$ is a small set and therefore we perform evaluations of Equations \ref{eq:f} and \ref{eq:pi} for the whole set at each step of the game. In our second experiment, the Hanabi game, $F^{S,s_t}$, which is the set of all possible hands that $A^S$  can be privy to, is large and we sample from it at each step of the game.

We further assume that self-play is the optimal response policy in our experiments and therefore $B(\hat{\pi}_{t}^S) = \hat{\pi}_{t}^S$. The reason we choose self-play as an optimal response policy is that a player's best response has to be a strategy that maximizes its expected reward and ideally will result in a Nash equilibrium, where no other player has an incentive to play a different strategy. As discussed earlier, in a two-player setting where regret is minimised, although not guaranteed a self-play strategy could still converge to a Nash equilibrium and perform very well in non zero-sum-games \cite{gibson2013regret}. We discuss this further in Section \ref{sec:discussion}.

\subsection{Slime Volleyball}

\begin{wrapfigure}{R}{0.4\textwidth}
    \vspace{0mm}
    \centering
    \includegraphics[width=0.35\textwidth]{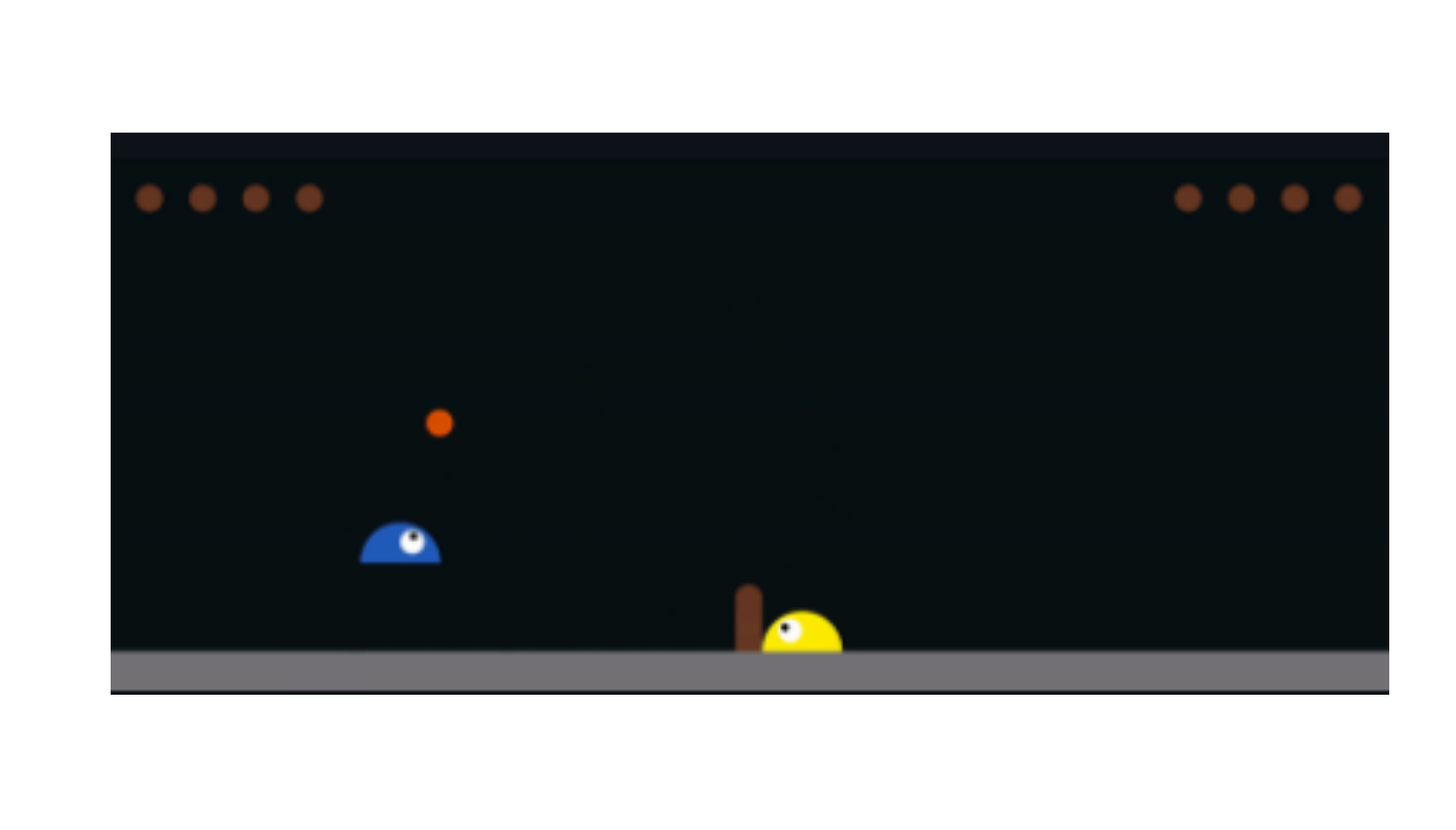}
    \vspace{0mm}
    \caption{\small \emph{Slime Volleyball}.}
    \label{fig:slimevolley}
    \vspace{0mm}
\end{wrapfigure}\vspace*{0mm}

We begin by considering the game of volleyball. We modified the recently introduced \emph{Slime Volleyball} (slimevolleygym, \cite{slimevolleygym}) such that the two ``slime'' agents play together in a cooperative fashion, seeking to keep the ball in the air for as long as possible (see Fig. \ref{fig:slimevolley}). In addition, we alter the agent observation to make it partially observable, inducing a coordination problem as the agents cannot see the horizontal ball speed when the ball is in their side of the court, thus must infer this information along with the strategy of the other agent.

We consider three types of agents, one trained in self-play mode with deep RL (PPO, \cite{schulman2017proximal}), one trained in self-play mode with a genetic algorithm (GA. \cite{genetic_algo}), and finally a scripted bot (Bot) which takes an action based on simple set of rules.

We assess the performance using two indicators of value, namely the average duration of each game (time) and the total number of passes between the players. Higher values for these indicators imply the agents are coordinating effectively.

\Heatset{min=0,   
           max=25,   
           max colour=Burlywood2, 
           min colour=white,      
           Min colour=black, 
}


\begin{table}[H]
\vspace{0mm}
\centering\begin{adjustwidth}{0cm}{}
\emph{{Slime volleyball cross-play with and without OSA. Each cell shows the mean $\pm$ standard error for time and number of passes for 1000 games.  The first row in each cell is the duration time and the second the number of passes. Larger values imply improved coordination performance.\\}}\vspace*{0.2cm}
\scalebox{0.79}{

\parbox{.45\linewidth}{
\caption{Without OSA}\label{volley_1}
\begin{tabular}{c c c c c}

\multicolumn{1}{c}{}&\multicolumn{1}{c}{}&\multicolumn{3}{c}{$\pi^S$ } \\
\multicolumn{1}{c}{}& &Bot & GA & PPO \\ 
\cline{3-5}           
\multirow{3}{*} {$\pi^C$}&Bot & \makecell{$2654 \pm15 $ \\ $ 24.44\pm0.19 $} & \makecell{ $2285 \pm19 $ \\$ 21.24\pm0.22 $ } &  \makecell{$ 2200\pm 19$ \\$23.58  \pm0.22 $}\\  
& GA & \makecell{ $2268 \pm20 $ \\ $22.82 \pm0.21 $} & \makecell{ $ 2724\pm14 $ \\$ 32.23\pm0.21 $} & \makecell{ $ 1259\pm12 $ \\ $13.37 \pm0.19 $}  \\ 
 &PPO &\makecell{$2222 \pm 19$  \\ $23.99 \pm 0.25$} & \makecell{ $1254 \pm 12$ \\ $15.22 \pm0.15 $} & \makecell{ $ 1979\pm18 $ \\$ 26.79\pm0.32 $}  \\ 
 \cline{3-5}           

\end{tabular}
}

\hspace{2.5cm}

\parbox{.45\linewidth}{
  \caption{\small{With OSA}}\label{volley_2}
\begin{tabular}{c c c c c}
\multicolumn{1}{c}{}&\multicolumn{1}{c}{}&\multicolumn{3}{c}{$\pi^S$} \\
  
\multicolumn{1}{c}{}& &Bot & GA & PPO \\ 
\cline{3-5}         
\multirow{3}{*} {$\pi^C_0$}& Bot& \makecell{$2642 \pm16 $ \\ $24.41  \pm 0.19$ } & \makecell{ $ 2730\pm 13$ \\ $ 32.08 \pm0.20 $  } &  \makecell{ $ 2000\pm18 $ \\ $27.44 \pm 0.34$}\\  
& GA& \makecell{ $2677 \pm15 $ \\$ 24.84\pm0.19 $} & \makecell{  $2737 \pm13 $ \\ $ 32.47\pm0.20 $} & \makecell{ $ 1950\pm19 $  \\$26.35 \pm0.35 $}  \\ 
 &PPO&\makecell{ $ 2645\pm 16$ \\$ 24.68\pm 0.19$} & \makecell{ $ 2704\pm14 $ \\ $32.20 \pm0.21 $ } & \makecell{$ 1965\pm18 $  \\$ 26.72\pm 0.34$}  \\ 
 \cline{3-5}         
\end{tabular}
}
}
\end{adjustwidth}
\vspace{0mm}
\end{table}

In Table \ref{volley_1} we show the cross-play matrix for the three agents in our set. For example, the PPO agent can play for on average 26.79 passes, and the GA agent can play for 32.23 passes - yet when they are paired together the numbers are significantly lower (15.22 and 13.37, with numbers differing due to the ordering of the agents). When we use OSA to adapt the policy from the entire set, we are able to achieve a significantly better return (Table \ref{volley_2}). Not only that OSA outperforms the cross-play matrix, but also for each $\pi^S$, the OSA is able to achieve the self-play performance levels.

For the number of passes the results achieved by OSA are higher in all cases and statistically significant using the standard errors. For the average time, OSA achieves levels that are higher and statistically significant as well, with the exception of $\pi^S$ playing the PPO policy and the $\pi^C_0$ the Bot policy. The number of passes are still higher in this case when using OSA. The reason for the higher average time for the cross-play compared to OSA is that the Bot policy is a much slower strategy compared to the PPO and the Bot agent keeps the ball in his court for a longer duration compared to the PPO agent. So while the average time is higher for the cross-play without OSA, the agents cooperation in passing the ball to each other is less.

\subsection{Hanabi Experiments}

We now consider ad-hoc coordination in the challenging Hanabi learning environment \cite{bard2020hanabi}. A brief description of the game and rules are as follows. There are five cards in each player's hand. There are five colors; for each color, card values are as follows: one 5, three 1s, and two cards of ranks 2,3 \& 4, providing 10 cards per color and a total of 50 cards in a deck. The players can observe the hands for all other players but not their own. The players need to cooperate by exchanging information using hints. The latter can help players make a decision as to which card to play. The goal of the game is to make a `firework' for each color, consisting of all the card ranks of that color. Five fireworks can hence be made and the maximum score for each game is 25. There are 20 actions possible for each player at each turn; a player can either play or discard one of the 5 cards they are holding or they can hint to the rank or the color of the other players' holdings. The game begins with eight available information tokens and three life tokens. The game ends immediately if all life tokens are gone. The game can also come to an end if all fireworks have been successfully made, or after a player draws the last card from the deck and every player has taken one final turn. 

Moving to ad-hoc coordination, a wide range of agents have been considered for Hanabi. In this paper we consider coordination amongst the following: Rainbow \cite{rainbow}, hand coded bots from the Hanabi Open Agent Dataset (HOAD, \cite{hoad}), Valuebot, Holmesbot, Iggi and Piers, and Multi Agent PPO (MAPPO, \cite{mappo}). Notably, this includes both on policy (MAPPO) and off policy (Rainbow) deep RL approaches, as well as scripted bots, providing a diverse range of agents. For the MAPPO model, we use two separate trained policies with different seeds. 

With regards to the bot strategies the Valuebot plays cards that it has enough knowledge to know that they are playable, and gives hints about the cards that are valuable. It also prefers to discard the oldest cards first. Holmesbot is similar to Valuebot, but uses the life tokens and additional inference to strategize \cite{hoad}. Iggi plays cards that it has enough information about as well and prefers number hints over color hints. Unlike the Valuebot it prefers discarding unplayable cards first and before the oldest card. Piers is similar to Iggi but also uses the life tokens and additional rules in order to strategize and avoid discarding the cards that are valuable \cite{hoad, canaan2020evaluating}. Therefore there is a positive correlation between Valuebot and Holmesbot strategies on one hand, and Iggi and Piers strategies on the other hand.

\subsubsection{Using OSA for Hanabi}

We focus on a 2-player version of Hanabi, consisting of agents $A^C$ and $A^S$. The hidden information from $A^C$ at each step is its own hand $H^C_t$, which is known by $A^S$. Therefore $A^C$ attempts to estimate the joint probability distribution $P(\pi^S, H^C_t \mid u^S_t, s_t)$ at each step using OSA (Algorithm \ref{osa_algo}), where $f^{S,s_t} = H^{C,s_t}$. We now evaluate the performance of agents in a variety of settings, with and without OSA. All evaluations are averaged over 1000 games.

\subsubsection{Ad-hoc team play}

As with the slime volleyball setting, we begin by considering the cross-play matrix of the agents in our set, which we show in Table \ref{hanabi_include_xp}. As can be seen, compatibility is non-trivial, for example Rainbow achieves 22.01 in self play, but less than one point with Holmesbot or Valuebot. In addition, even the same algorithm may not repeatedly discover compatible strategies, the MAPPO agents both perform drastically worse together than when partnered with a copy of themselves. 

To test OSA, we conduct two sets of Hanabi experiments. In the first Hanabi experiment agent $A^S$'s policy is included in the $A^C$'s policy set, therefore $\pi^S \in \Pi$. In the first experiment we have a fixed set policies for all experiments. The results are shown in Table \ref{hanabi_include_osa}. As we see, when the agent is included in the set, OSA is able to recover performance close to self-play. Even in challenging settings, for example being initialized with a Holmesbot agent and playing with Rainbow, the algorithm still achieves a score of 19.54. 

Next we consider a more challenging setting, where $A^C$ does not have $A^S$'s policy, $\pi^S \not \in \Pi$. In this second experiment, for each evaluation we exclude $\pi^S$ from our set. The results are shown in Table \ref{hanabi_exclude_osa}. As we see, despite not containing the optimal policy, $A^C$ still successfully plays with $A^S$ using the policy in its set that is most correlated with $\pi^S$ for majority of the games. Comparisons for both experiments, which both contain statistically significant gains are shown in Table \ref{zero_shot_comp}.

\Heatset{min=0,   
           max=25,   
           max colour=Burlywood2,
           min colour=white,      
           Min colour=black, 
}

\begin{table*}
\begin{adjustwidth}{-0cm}{}
{\emph{Mean rewards for ad-hoc play between agents $A^C$ and $A^S$, when the simple agent's policy is part of the complex agent's policy set.}\\ }
\scalebox{0.61}{
\parbox{.8\linewidth}{
\caption{Without OSA}\label{hanabi_include_xp}

  \hspace*{-2em}\begin{tabular}{ cc HHHHHHHH }
          & & \multicolumn6c{\textsf{\small $\pi^s$}} & \\
         &  &\multicolumn1c{\textsf{\small [i]}}&\multicolumn1c{\textsf{\small [ii]}}
            &\multicolumn1c{\textsf{\small [iii]}}&\multicolumn1c{\textsf{[iv]}} &\multicolumn1c{\textsf{\small [v]}} & \multicolumn1c{\textsf{\small [vi]}} &  \multicolumn1c{\textsf{\small [vii]}}\\
            
          & \textsf{\small [i] MAPPO-1}&23.99    &15.22  &0.00  &3.91   &8.87  	 &9.45   &0.08  \\
         & \textsf{\small [ii] MAPPO-2}&16.41   &23.41  &0.00  &8.16   &8.05  	 &16.3   & 0.06 \\
        
             \hspace*{-5mm}\rotatebox{0}{\makebox[-5pt]{\textsf{\small $\pi^C$}}}\hspace*{-3mm}
         & \textsf{\small [iii] Holmesbot}& 0.00   &0.02  &15.76  &1.96   &0.32  	 &0.19   & 17.44 \\
         & \textsf{\small [iv] Iggi}&4.50    &7.96  &2.06  &16.97   &12.75 &5.17   & 4.21 \\
         & \textsf{\small [v] Piers}&9.01    &8.41  &0.37  &13.24   &17.17 &8.64   &1.17  \\
         & \textsf{\small [vi] Rainbow}&8.90    &15.66  &0.25  &4.56   &8.58 &22.01   & 0.63 \\
         & \textsf{\small [vii] Valuebot}&0.04    &0.06  &17.49  &3.58   & 0.99 	 & 0.39  & 18.5 \\

  \end{tabular}
  }
 \hspace{-0.4cm}
  \parbox{.8\linewidth}{
  \caption{With OSA}\label{hanabi_include_osa}
  
  \begin{tabular}{ cc HHHHHHHH }
           &  \multicolumn7c{\textsf{\small \quad\quad\quad $\pi^S$}} & \\
         &  &\multicolumn1c{\textsf{\small [i]}}&\multicolumn1c{\textsf{\small [ii]}}
            &\multicolumn1c{\textsf{\small [iii]}}&\multicolumn1c{\textsf{\small [iv]}} &\multicolumn1c{\textsf{\small [v]}} & \multicolumn1c{\textsf{\small [vi]}} &  \multicolumn1c{\textsf{\small [vii]}}\\
          & \textsf{\small [i] MAPPO-1}&23.21  &21.72    & 14.06 & 15.99  &  16.47	 &20.08   & 16.13 \\
         & \textsf{\small [ii] MAPPO-2}& 23.07 &22.07  &13.60  &15.96    &  	16.51 & 20.12  & 15.71 \\
             \hspace*{-5mm}\rotatebox{0}{\makebox[-5pt]{\textsf{\small $\pi^C_0$}}}\hspace*{-3mm}
         & \textsf{\small [iii] Holmesbot}&22.72  &21.05  & 14.00  &16.02   &  16.29	 &19.54   & 16.51 \\
         & \textsf{\small [iv] Iggi}&23.05  &21.89  &13.35  &16.14     &  	 16.43&20.20   & 16.12 \\
         & \textsf{\small [v] Piers}&23.03  &22.03  &13.62  &16.30     &  	 16.30&19.92   & 15.92 \\
         & \textsf{\small [vi] Rainbow}&22.97  &21.93  & 13.24   &16.16   &  16.24	 &20.16   &  15.90\\
         & \textsf{\small [vii] Valuebot}&22.70  &21.47  &14.67    &16.04   &  16.23	 &19.83   & 16.43 \\

  \end{tabular}
  }
  }
  \end{adjustwidth}
 \vspace{0mm}
\end{table*}
\begin{table*}
\begin{adjustwidth}{-0cm}{}
{\emph{Mean rewards for ad-hoc play between agents $A^C$ and $A^S$, when the simple agent's policy \textbf{is excluded} from the complex agent's policy set.\\}}
\scalebox{0.61}{
\parbox{.8\linewidth}{
  \caption{Without OSA}\label{hanabi_exclude_xp}
  \hspace*{-2em}\begin{tabular}{ cc HHHHHHHH }
          & & \multicolumn6c{\textsf{\small $\pi^S$}} & \\
         &  &\multicolumn1c{\textsf{\small [i]}}&\multicolumn1c{\textsf{\small [ii]}}
            &\multicolumn1c{\textsf{\small [iii]}}&\multicolumn1c{\textsf{\small [iv]}} &\multicolumn1c{\textsf{\small [v]}} & \multicolumn1c{\textsf{\small [vi]}} &  \multicolumn1c{\textsf{\small [vii] }}\\
          & \textsf{\small [i] MAPPO-1}&-1   &15.22  &0.00  &3.91   &8.87  	 &9.45   &0.08  \\
         & \textsf{\small [ii] MAPPO-2}&16.41   &-1  &0.00  &8.16   &8.05  	 &16.3   & 0.06 \\
             \hspace*{-5mm}\rotatebox{0}{\makebox[-5pt]{\textsf{\small $\pi^C$}}}\hspace*{-3mm}
         & \textsf{\small [iii] Holmesbot}& 0.00   &0.02  &-1  &1.96   &0.32  	 &0.19   & 17.44 \\
         & \textsf{\small [iv] Iggi}&4.50    &7.96  &2.06  &-1   &12.75 &5.17   & 4.21 \\
         & \textsf{\small [v] Piers}&9.01    &8.41  &0.37  &13.24   &-1 &8.64   &1.17  \\
         & \textsf{\small [vi] Rainbow}&8.90    &15.66  &0.25  &4.56   &8.58 &-1   & 0.63 \\
         & \textsf{\small [vii] Valuebot}&0.04    &0.06  &17.49  &3.58   & 0.99 	 & 0.39  & -1 \\

  \end{tabular}
  }
  \hspace{-0.2cm}
  \parbox{.8\linewidth}{
  \caption{With OSA}\label{hanabi_exclude_osa}
  \begin{tabular}{ cc HHHHHHHH }
          &  \multicolumn7c{\textsf{\small \quad\quad\quad\quad\quad $\pi^S$}} & \\
         &  &\multicolumn1c{\textsf{\small [i]}}&\multicolumn1c{\textsf{\small [ii]}}
            &\multicolumn1c{\textsf{\small [iii]}}&\multicolumn1c{\textsf{\small [iv]}} &\multicolumn1c{\textsf{\small [v]}} & \multicolumn1c{\textsf{\small [vi]}} &  \multicolumn1c{\textsf{\small [vii]}}\\
          & \textsf{\small [i] MAPPO-1}& -1 & 12.37   & 13.11 & 9.83  & 9.52 	 &  11.28 & 13.85 \\
         & \textsf{\small [ii] MAPPO-2}&10.74  & -1 & 13.31 & 10.18   & 9.90 	 &   11.44& 14.02 \\
             \hspace*{-5mm}\rotatebox{0}{\makebox[-5pt]{\textsf{\small $\pi^C_0$}}}\hspace*{-3mm}
         & \textsf{\small [iii] Holmesbot}& 10.20 & 10.80 &  -1 &10.01 &  9.62	 & 11.07  & 14.58 \\
         & \textsf{\small [iv] Iggi}&10.94  & 12.45 & 13.19 &  -1   & 10.02 	 &  11.88 &  14.52\\
         & \textsf{\small [v] Piers}& 10.26 & 12.12 &13.33  &9.92     &  -1	 & 11.54  & 14.33 \\
         & \textsf{\small [vi] Rainbow}&10.57  &12.35  & 13.26   &10.38   & 10.16  	 &  -1 &  14.36\\
         & \textsf{\small [vii] Valuebot}&9.95  & 11.29 &13.65    &9.87   & 10.28 	 & 10.94  & -1 \\

  \end{tabular}
  }
  }
  \end{adjustwidth}
\vspace{0mm}
\end{table*}

\begin{table}[ht]

\begin{adjustwidth}{-0.2cm}{}
\caption{\small{Mean $\pm$ std err of rewards for $A^C$ playing each of the 7 policies in the two Hanabi experiments.} }\label{zero_shot_comp}
\vspace{0mm}
\scalebox{0.75}{
\begin{tabular}[t]{lccccccc}
\toprule
&{MAPPO-1} &{MAPPO-2}  &{Holmesbot} &{Iggi} &{Piers} &{Rainbow} &{Valuebot} \\

\midrule
$\mathbf{\pi^S \in \Pi :} $ &&&&&&& \\
$\mathbb{E}_\tau R(\tau)$ no OSA  & $ 8.98\pm 0.12 $& $ 10.15 \pm 0.12 $ &$5.13 \pm 0.10 $ & $ 7.48\pm0.09 $ &$8.10 \pm0.10  $&$8.87 \pm0.11 $& $6.01 \pm0.10 $\\

$\mathbb{E}_\tau R(\tau)$ with OSA & $22.96 \pm0.05 $& $21.74 \pm0.06 $ &$13.79 \pm0.10 $ & $ 16.09\pm0.04 $ &$16.35 \pm0.04 $ & $19.98 \pm0.07 $& $ 16.10\pm0.07 $\\

&&&&&&& \\
$\mathbf{\pi^S \not\in \Pi :} $&&&&&&& \\
$\mathbb{E}_\tau R(\tau)$ no OSA  & $6.47 \pm 0.11 $& $ 7.94 \pm0.12  $ &$3.36 \pm 0.09 $ & $5.90 \pm0.09 $  &$6.59 \pm0.10  $ & $6.69 \pm 0.10$& $3.93 \pm 0.09$\\

$\mathbb{E}_\tau R(\tau)$ with OSA &$10.44\pm0.12 $&$11.90 \pm0.13 $&$ 13.31\pm0.10 $&$10.03 \pm0.10 $&$ 9.92\pm0.11 $&$ 11.36\pm0.11 $&$14.28 \pm0.09 $ \\

\bottomrule
\end{tabular}
}
\end{adjustwidth}
\vspace{0mm}
\end{table}

\subsubsection{Complex agent policy distribution in ad-hoc games}

\begin{figure}[h!]
\vspace{0mm}
    \hspace{0cm}
    \centering
    \begin{minipage}[c]{0.85\textwidth} 
     \includegraphics[width=.99\linewidth]{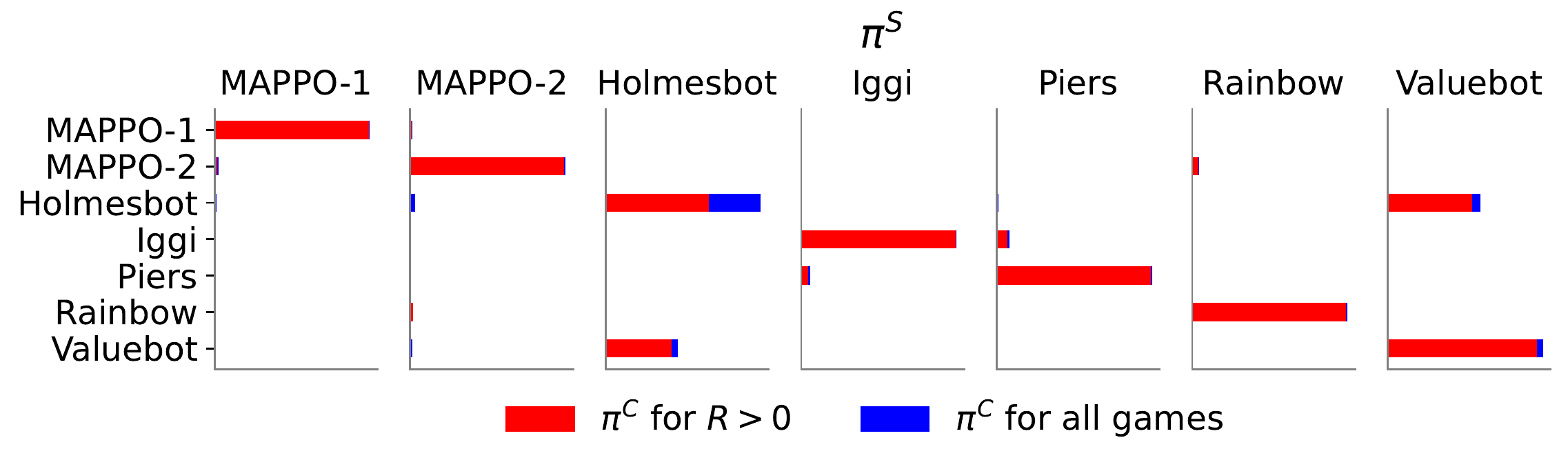}
    \end{minipage}
    \vspace{0mm}
    \caption{\small{Distribution of $\pi^C_T$ for each $\pi^S$ in the first Hanabi experiment in which $\pi^S \in \Pi$.}}
    \label{fig:include_dis}
    \vspace{0mm}
    
\end{figure}

In order to better understand the strength and weakness of the method, we assess the distribution of the policies that $A^C$ plays in coordination with $A^S$ for each Hanabi experiment. Figure \ref{fig:include_dis} depicts the distribution for $\pi^C_T$ in the first Hanabi experiment ($\pi^S \in \Pi$) in which $\pi^C_T$ is agent $A^C$'s policy at the end of each game. The blue bars indicate the distribution over all games, irrespective of reward values at the end of each game. The red bars indicate the distribution only for the games in which rewards were greater than zero. In the first experiment, with the exception of the Holmesbot and Valuebot strategies, agent $A^C$ has close to perfect accuracy in identifying $\pi^S$ and playing the policy with $A^S$. In the case of Holmesbot and Valuebot strategies, the two policies have a high correlation and excellent cross-play scores. Whilst OSA identifies $\pi^S$ accurately in majority of the games, at times it identifies Holmesbot as Valuebot and vice versa. We note, however, that for these two policies the games that are wrongly identified as the other type still yield a positive reward with very few exceptions.

\begin{figure}[h!]

    \centering
    \begin{minipage}[c]{0.85\textwidth} 
     \includegraphics[width=.99\linewidth]{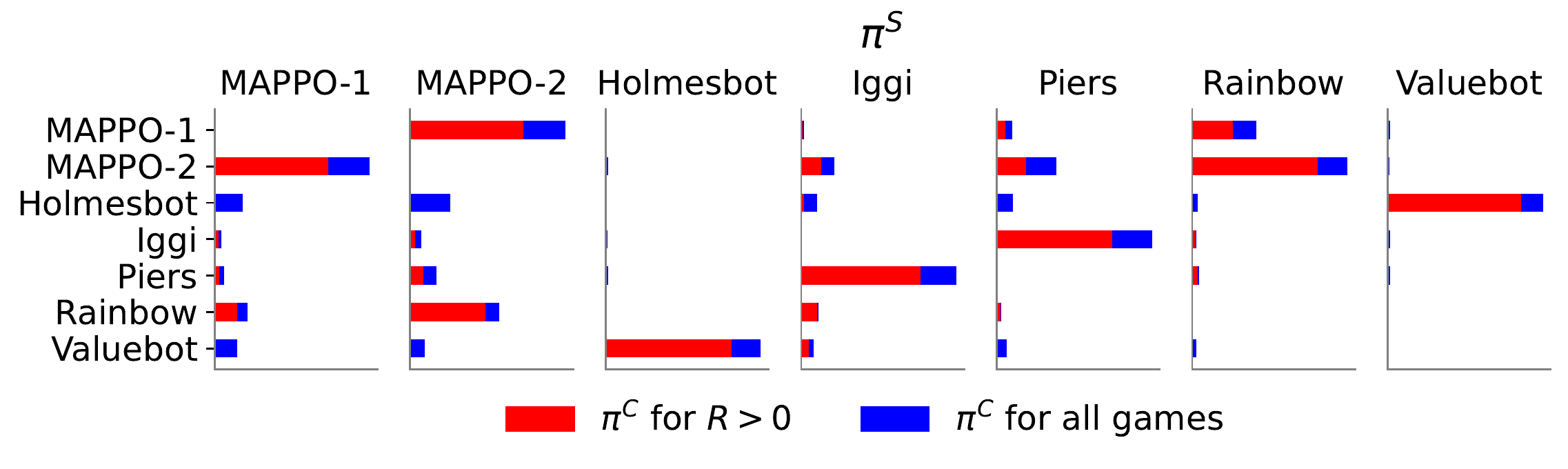}
    \end{minipage}
    \vspace{0mm}
    \caption{\small{Distribution of $\pi^C_T$ for each $\pi^S$ in the second Hanabi experiment where, $\pi^S \not \in \Pi$.}}
    \label{fig:exclude_dis}
    \vspace{0mm}
\end{figure}

Figure \ref{fig:exclude_dis} depicts the distribution of $\pi^C_T$ in the second Hanabi experiment, in which $\pi^S \not \in \Pi$. In this experiment, across all seven policies, agent $A^C$ plays the policy with the highest cross-play score with $\pi^S$ for the majority of games. In this experiment we note that $A^C$ also plays, on occasion, other policies that have a correlation with $\pi^S$. We further note that in many games that achieved zero scores, $\pi^C_T$ was still the policy with the closest cross-play score to $\pi^S$.

In summary, in the first Hanabi experiment, $\pi^S$ is the dominant policy selected by $A^C$ to play $A^S$, and in the second Hanabi experiment, where $\pi^S \not \in \Pi$, the policy with the highest cross-play score with $\pi^S$ is the dominant strategy for $A^C$ to play with $A^S$.

\subsubsection{$k$-shot coordination}

All experiments performed in the previous section are single-shot ad-hoc coordination games in which an agent $A^C$ has no knowledge of agent $A^S$'s policy, and has not played with $A^S$ previously. In this section we define a $k$-shot ad-hoc game whereby $A^C$ plays with $A^S$ a total of $k$ times in the second Hanabi game ($\pi^S \not \in \Pi$). We note that, in the first Hanabi experiment, the agents already achieve close to self-play performance levels in a single-shot ad-hoc setting.

For the $k$-shot games, agent $A^C$ begins the first game by using OSA and continues until a positive reward is achieved. If the agents achieve a positive reward in the $m^{\text{th}}$ game of the $k$ games, $A^C$ uses $\pi^{C,m}_T$ for the subsequent games, where $\pi^{C,m}_T$ is $A^C$'s last step policy of the $m^{\text{th}}$ game using the OSA method. In short $A^C$ plays as follows:
\begin{equation}
    \pi^{C,m}=
    \begin{cases}
      \pi^{C,m-n}_T, & \text{if}\ (R^{m-n}>0) \land (m > n\geq 0)\\
      \pi^{C,m}_t, & \text{otherwise}
    \end{cases}
  \end{equation}
where $R^{m-n}$ is the reward for game $m-n$. Table \ref{k_shot} shows the results for $k$-shot games, where $k \in \{0, 1,4\}$, in the second experiment, in which $\pi^S \not \in \Pi$. These results are  compared with the maximum cross-play scores of $A^C$ policies against $\pi^S$. In the 4-shot game, for all policies, the agents achieve rewards close to this maximum level.

\begin{table}[ht]
\begin{adjustwidth}{-0.3cm}{}
\caption{\small{Mean $\pm$ std err of rewards in $k$-shot games. $A^C$ plays each of the 7 policies using OSA and $\pi^S \not\in \Pi $.}}\label{k_shot}
\vspace{0mm}
\scalebox{0.75}{
\begin{tabular}[t]{lccccccc}
\toprule
&{MAPPO-1} &{MAPPO-2}  &{Holmesbot} &{Iggi} &{Piers} &{Rainbow} &{Valuebot} \\

\midrule
$\mathbb{E}_\tau (R(\tau) \mid k):$&&&&&&& \\
$\quad\quad k=0$ &$10.44\pm0.12 $&$11.90 \pm0.13 $&$ 13.31\pm0.10 $&$10.03 \pm0.10 $&$ 9.92\pm0.11 $&$ 11.36\pm0.11 $&$14.28 \pm0.09 $ \\
$\quad\quad k=1$  &$12.84\pm0.12$&$ 13.76\pm0.12 $&$16.45 \pm0.08 $&$10.76 \pm0.10 $&$10.86 \pm0.11 $&$13.45 \pm0.11 $&$ 16.72\pm0.08 $ \\
$\quad\quad k=4$   &$14.52\pm 0.11$&$14.86 \pm0.11 $&$ 17.33\pm0.07 $&$ 11.05\pm0.10 $&$ 11.43\pm0.10 $&$14.21 \pm0.10 $&$ 17.24\pm0.07 $ \\
&&&&&&& \\
$\max_{i=0}^n (\mathbb{E}_\tau R_i)$ &$16.41\pm 0.11$ &$ 15.66\pm0.11 $&$17.49 \pm 0.07 $&$ 13.24\pm0.09 $&$12.75 \pm0.09 $&$16.30 \pm0.10 $&$ 17.44\pm0.07 $ \\

\bottomrule
\end{tabular}
}
\end{adjustwidth}
\vspace{0mm}
\end{table}%

\subsubsection{Comparison with a Common Best Response to a Population Policy}

In this section we compare the performance of the OSA algorithm with a common best response (CBR) to a population of agents policy. The CBR policy trains on all seven strategies using the MAPPO optimisation method. Table \ref{BR_table} shows the performance of this policy compared to the OSA algorithm in a single-shot game for both experiments. Both OSA experiments perform significantly better compared to the CBR Policy.

\begin{table}[ht]
\begin{adjustwidth}{-0.3cm}{}
\caption{\small Mean $\pm$ standard error of rewards for the CBR Policy and the two OSA experiments.}\label{BR_table}
\vspace{0mm}
\scalebox{0.75}{
\begin{tabular}[t]{lccccccc}
\toprule
&{MAPPO-1} &{MAPPO-2}  &{Holmesbot} &{Iggi} &{Piers} &{Rainbow} &{Valuebot} \\

\midrule

CBR Policy &$5.36\pm0.12 $&$6.03 \pm0.13 $&$ 1.11\pm0.07 $&$2.38 \pm0.08 $&$ 3.67\pm0.11 $&$ 4.28\pm0.12 $&$2.20 \pm0.06 $ \\

OSA ($\pi^S \not \in \Pi$)  &$10.44\pm0.12 $&$11.90 \pm0.13 $&$ 13.31\pm0.10 $&$10.03 \pm0.10 $&$ 9.92\pm0.11 $&$ 11.36\pm0.11 $&$14.28 \pm0.09 $ \\

OSA ($\pi^S \in \Pi$)   & $22.96 \pm0.05 $& $21.74 \pm0.06 $ &$13.79 \pm0.10 $ & $ 16.09\pm0.04 $ &$16.35 \pm0.04 $ & $19.98 \pm0.07 $& $ 16.10\pm0.07 $\\

\bottomrule
\end{tabular}
}
\end{adjustwidth}
\vspace{0mm}
\end{table}%

\subsection{Discussion}\label{sec:discussion}

In our Hanabi experiments we have included a set of policies from a diversified set of models. We note that agent $A^C$'s policy set, $\Pi$, is not restrictive and can include other models. For example the strategies presented in \cite{hu2020other} and \cite{offbelieflearning} are shown to play the Hanabi game well with humans in ZSC settings; these policies could be added to agent $A^C$'s policy set. The OSA algorithm is a method that is inclusive of all strategies and models and not in direct comparison with each. It may also be possible to achieve strong ad-hoc coordination by \emph{learning} a set of diverse strategies and then using OSA at test time, e.g. via Fictitious Co-Play \cite{strouse2021fcp}. 

Further, due to the policy-removal addition to the Gibbs sampling algorithm (steps 5 to 9 in Algorithm \ref{osa_algo}), the method is able to focus the policy set to the most relevant subset early on in the game, thereby scaling well as the number of policies in $\Pi$ are increased. In addition, performance is further improved in $k$-shot games with the OSA method and agents can achieve performance close to the best cross-play in the policy set, when $\pi^S$ is not in $\Pi$. We note that this improvement can not be achieved for strategies that play a fixed policy in every play of the $k$-shot game. 

Regarding the limitations of OSA, $A^C$'s performance is dependant on the policies in $\Pi$ and $\pi^S$. If none of the policies in $\Pi$ are compatible with $\pi^S$, OSA will not perform well. A well diversified portfolio of policies will mitigate this issue to some extent. This further implies that if $\pi^S$ is a random play, the algorithm will not have any added value. We further note that, in the current format, the method treats all policies in the policy set $\Pi$ equally and there are no portfolio optimization steps for policy selection. We see this as a possible extension.

We have not investigated in detail the alternative approach of training a best response policy in this work. The OSA algorithm performs two main steps. In the first step the algorithm evaluates the other agents by estimating the joint probability, $P(\pi^S,f^{S,s_t} \mid u_t^S, s_t)$, discussed in Section \ref{sec:Gibbs_sampling}. The second step is to determine the best response policy after this evaluation. We chose self-play in our experiments. However it is possible that self-play is not the optimal response policy for certain MARL settings. While there is a lot of empirical evidence for the success of self-play strategies, more theoretical work is needed to better understand their effectiveness specifically in non zero-sum-games. Even though the two-player self-play game could converge to a Nash equilibrium this is not a guaranteed outcome in non zero-sum-games specially in games with partial observability. We also note that finding a Nash equilibrium itself is a minimal requirement for a satisfactory result in a multi-agent system. While it could be the only optimal solution in some games, it is possible to have multiple Nash equilibria and in that case for a good performance it is reasonable to require the game to converge to a Pareto optimal Nash equilibrium \cite{conitzer2007awesome}, as in this case simply achieving a Nash equilibrium could be sub-optimal. Further in our experiments the games are symmetric with respect to the agents and no agent has an advantage compared to others. Choosing self-play as a best response policy in an environment where agents have non-symmetric resources could lead to an undesirable outcome.

\section{Conclusion and Future Work} \label{sec:conclusion}

We presented On-the-fly Strategy Adaptation (OSA), a novel approach for coordination between ad-hoc agents across a set of diverse models. Despite its simplicity, the algorithm  achieves impressive performance while scaling gracefully. We have further provided average rewards for both ad-hoc and $k$-shot ad-hoc games and shown that performance improves in $k$-shot games. We believe OSA could be particularly impactful in partially observable cooperative settings, which feature in many real world settings.

Considering potential extensions to improve performance, we note that the complex agent's choice of its policy set, $\Pi$, can be optimized. In our current setting, policies are treated equally and no analysis is made of the models that generate these policies, nor their policy cross-correlations. We note that there is such cross-correlation amongst policies in most games. A potential extension to OSA could consider moving beyond simply removing policies and allowing policy creation and evolution, similar to reversible jump Markov chain Monte Carlo methods \cite{roberts2001minimum}. This may allow for a dynamic policy set at each step of the game.

Potential extensions to the algorithm could re-evaluate the approach for the best response policy and to consider policies that are not self-play. Further work can consider using the OSA approach for strategy selection alongside a method for generating diverse strategies \cite{mapelites_hanabi, determinantalqlearning, trajedi, dvd, novelty}. Furthermore, it may even be possible to directly optimize agents to achieve strong ad-hoc coordination via our approach. For example, we could train agents in such a fashion that they provide useful contributions to the policy set. Other extensions of the work offers relaxation of the assumption that agents $A^S$ play a fixed policy throughout the game. It is reasonable to assume that some agents in the system could play a mixed strategy during the game.

Finally, we note that in both of our experiments the rules of the game are fixed. In real-world problems, rules themselves are a variable within the system and subject to change. Since OSA uses the conditional probabilities to estimate the joint probability, in theory it is possible to adapt to evolving rules, though this is a particularly challenging problem.

\section*{Acknowledgments}

The authors would like to thank Jakob Foerster and anonymous reviewers for useful feedback that helped improve our work. The experiments in this paper were made possible thanks to AWS.

\bibliographystyle{abbrv}
\bibliography{refs}

\end{document}